%% file: iclr2025_conference.tex
\documentclass{article} 
\pdfoutput=1

\usepackage{iclr2025_conference}
\usepackage{times}

\makeatletter
\renewcommand{\@oddhead}{} 
\renewcommand{\@evenhead}{} 
\makeatother

\input{math_commands.tex}

\usepackage{hyperref}
\usepackage{url}
\usepackage{graphicx}
\usepackage{booktabs}

\title{Toward Trustworthy Difficulty Assessments: Large Language Models as Judges in Programming and Synthetic Tasks}

\author{
  \textbf{H.M. Shadman Tabib}$^{*}$ \quad 
  \textbf{Jaber Ahmed Deedar}$^{*}$ \\
  Department of Computer Science, Bangladesh University of Engineering and Technology \\
  \texttt{shadmantabib2002@gmail.com} \\
}

\iclrfinalcopy

\begin{document}
\maketitle

\begin{abstract}
Large Language Models (LLMs) have demonstrated impressive capabilities in natural language and code generation, and are increasingly deployed as \emph{automatic judges} of model outputs and learning activities. Yet, their behavior on structured tasks such as predicting the difficulty of competitive programming problems remains under-explored. We conduct a systematic comparison of GPT-4o, used purely as a natural-language difficulty assessor, against an interpretable LightGBM ensemble trained on explicit numeric and textual features. On a dataset of 1{,}825 LeetCode problems labeled Easy, Medium, or Hard, LightGBM attains 86\% accuracy, whereas GPT-4o reaches only 37.75\%. Detailed analyses, including confusion matrices and SHAP-based interpretability, show that numeric constraints---such as input size limits and acceptance rates---play a crucial role in separating Hard problems from easier ones. By contrast, GPT-4o often overlooks these cues and exhibits a strong bias toward simpler categories. We further probe GPT-4o through a synthetic Hard-problem generation protocol. Surprisingly, GPT-4o labels almost all of its own synthetic Hard problems as Medium, contradicting its tendency to downgrade real Hard problems to Easy. Our findings connect to recent work on LLMs-as-judges and automatic difficulty estimation in programming and education, and highlight concrete failure modes that must be addressed before LLM-based judges can be considered trustworthy in competitive programming, educational platforms, or reinforcement-learning pipelines.
\end{abstract}

\section{Introduction}
Large Language Models (LLMs), such as GPT-4-class models, are now central to many AI-driven applications in code generation, tutoring, and evaluation. LLMs trained on code and natural language can already solve non-trivial programming tasks \citep{chen2021evaluating,li2022alphacode}, and the emerging \emph{LLMs-as-judges} paradigm proposes to use them as automatic evaluators of model outputs, student answers, or even entire interactive pipelines \citep{li2024llmjudge,bavaresco2025llmsjudges,seo2025educationeval}. These developments build on a broader ecosystem of alignment and reward modeling methods, where learning from human feedback plays a crucial role \citep{christiano2017deep} and where language-based rationalizations are increasingly studied \citep{narang2020wt5}.

Competitive programming, however, poses a particularly challenging setting. Hard problems on platforms like LeetCode and Codeforces are designed to test deep algorithmic reasoning, sophisticated data structures, and careful handling of numeric constraints such as input sizes, time limits, and adversarial edge cases. Systems like AlphaCode and AlphaCode 2 illustrate that achieving competitive-level performance requires large-scale sampling and careful behavioral filtering, even for powerful transformer models \citep{li2022alphacode,alphacode22023}. Prior work in automatic difficulty estimation has shown that a mixture of content features and behavioral statistics can predict problem difficulty reasonably well \citep{zhao2018topicsdifficulty,effenberger2019difficulty,skarbalius2021difficulty,wang2024difficulty}, and that even simple complexity measures (e.g., input sizes, branching structure) often correlate with perceived difficulty \citep{effenberger2019difficulty,alkhuzaey2024review}.

In this work we ask: \emph{Can a state-of-the-art LLM, used as a black-box text-only judge, faithfully reproduce the Easy/Medium/Hard labels of competitive programming problems?} We study GPT-4o's behavior on a curated dataset of 1{,}825 LeetCode problems and compare it to an interpretable LightGBM ensemble \citep{ke2017lightgbm} that explicitly leverages numeric metadata (constraints, acceptance rates) alongside TF--IDF features. We then probe GPT-4o in a synthetic regime, where it is asked to generate and then self-label new Hard problems.

Our contributions are threefold:
\begin{enumerate}
    \item We provide a detailed comparison of GPT-4o and an interpretable gradient-boosted ensemble on LeetCode-style difficulty prediction, showing that the LLM underperforms a classical model by a large margin and exhibits strong bias toward easier labels.
    \item Through SHAP-based analysis \citep{lundberg2017shap}, we show that numeric constraints dominate the LightGBM ensemble's decision boundary for Hard problems, whereas GPT-4o appears to underweight or ignore such information.
    \item We reveal an inconsistency in GPT-4o's internal notion of ``Hard'' difficulty: real Hard problems are frequently downgraded to Easy, while self-generated Hard problems are almost uniformly labeled Medium. We situate these findings within the growing literature on LLMs-as-judges \citep{li2024llmjudge,bavaresco2025llmsjudges,seo2025educationeval} and automatic difficulty estimation \citep{intisar2018cluster,skarbalius2021difficulty,wang2024difficulty}.
\end{enumerate}

\section{Related Work}
\subsection{LLMs as Judges and Automatic Evaluators}
The idea of using LLMs as \emph{judges}---that is, as automatic evaluators of texts, answers, or model outputs---has rapidly gained traction. The survey of \citet{li2024llmjudge} provides a comprehensive taxonomy of LLM-based evaluation methods, covering functionality, methodology, applications, and limitations, and documents widespread adoption in summarization, dialogue, and generation benchmarks. Recent empirical work has raised concerns about reliability and bias: \citet{bavaresco2025llmsjudges} introduce JUDGE-BENCH and show that LLMs' agreement with human annotations varies widely across tasks and models, while \citet{seo2025educationeval} systematically assess LLM-based evaluators in education and highlight inconsistencies in feedback accuracy and stability over time.

Beyond general-purpose evaluation, several studies investigate LLM-as-judge pipelines in specific domains, such as search and ranking or mathematical coherence in narratives \citep{baysan2025llmjudge,keith2025mathjudge}. These works underscore a key tension: LLM-based judges are attractive because they are flexible and cheap to deploy, but their biases and failure modes may be very different from those of human experts. Our work contributes to this line by examining a \emph{structured} domain---competitive programming difficulty---where numeric constraints and algorithmic structure matter strongly, and where mis-calibrated judgments have tangible downstream consequences.

\subsection{LLMs, Code Generation, and Competitive Programming}
LLMs trained on large corpora of source code have achieved strong performance on many programming benchmarks. \citet{chen2021evaluating} introduced Codex, demonstrating substantial gains on HumanEval and other code synthesis tasks. Since then, surveys and engineering reports have documented the capabilities and limitations of modern code LLMs \citep{jiang2022surveycode,idrisov2024programcode}. For competitive programming specifically, AlphaCode \citep{li2022alphacode} achieved an average ranking in the top 54.3\% of human participants on Codeforces contests, while the AlphaCode~2 technical report \citep{alphacode22023} shows further gains powered by larger models and extensive sampling and filtering.

These systems, however, are primarily evaluated on \emph{solution quality}, not on their ability to \emph{judge} problem difficulty. In programming education, difficulty is known to interact with learner behavior and cognitive load \citep{wang2023difficultyorder}, and recent work has started to ask whether LLMs can serve as proxies for human item- or puzzle-difficulty judgments \citep{kreveld2015puzzles}. Our study focuses specifically on difficulty labels (Easy/Medium/Hard) in a LeetCode-like setting, and compares LLM-based judgments against an interpretable, numeric-feature-aware model.

\subsection{Automatic Difficulty Estimation in Programming and Education}
There is a substantial body of work on estimating difficulty for educational questions, puzzles, and programming exercises. In text-based question difficulty prediction, \citet{alkhuzaey2024review} survey automatic approaches and highlight the role of linguistic and structural features. For programming, prior work has used topic models and performance data from online judges to infer difficulty levels and latent topics \citep{zhao2018topicsdifficulty,yera2017recommendation}. \citet{intisar2018cluster} use cluster analysis to group programming problems by difficulty, while \citet{effenberger2019difficulty} and follow-up work \citep{pelanek2020classification} analyze the relationship between complexity measures and difficulty in introductory programming tasks.

More recently, neural approaches have emerged. \citet{skarbalius2021difficulty} propose an automatic programming problem difficulty evaluator based on text classification, and \citet{wang2024difficulty} introduce a multimodal difficulty model that combines textual descriptions with example solutions via pre-trained language models. These methods generally treat difficulty estimation as a supervised or semi-supervised learning problem over carefully engineered feature spaces.

Our work aligns with this literature but differs in two ways: (i) we explicitly contrast an interpretable gradient-boosted ensemble with a black-box LLM judge, and (ii) we explore not only real problems but also synthetic Hard problems generated by the LLM itself, revealing a previously undocumented inconsistency in its internal difficulty boundary.

\subsection{Interpretable Machine Learning for Trustworthy Judges}
Interpretable models such as gradient-boosted decision trees have long been favored in settings where transparency is crucial. LightGBM \citep{ke2017lightgbm} provides a highly efficient implementation of gradient boosting, and in combination with post-hoc explanation techniques like SHAP \citep{lundberg2017shap}, it enables fine-grained analysis of feature contributions at both global and local levels. Parallel efforts in NLP have explored models that generate natural-language rationales alongside predictions \citep{narang2020wt5}, but such explanations are not guaranteed to faithfully reflect the underlying decision process.

In our setting, we treat the LightGBM ensemble as a structured baseline whose decision-making can be audited via SHAP, and contrast it with the opaque behavior of GPT-4o. Our findings suggest that even a relatively simple model with transparent feature attributions can provide a more trustworthy difficulty signal than a state-of-the-art LLM judge that ignores numeric constraints.

\section{Methodology}

\subsection{Dataset and Labeling}
We use a dataset of 1{,}825 LeetCode problems, each labeled by the platform as Easy, Medium, or Hard. For each problem we collect the full English description and two categories of metadata:

\begin{itemize}
    \item \textbf{Textual features}: TF--IDF vectors over unigrams and bigrams extracted from the problem statement, following common practice in text-based difficulty prediction \citep{zhao2018topicsdifficulty,alkhuzaey2024review}.
    \item \textbf{Numeric features}: scalar indicators including maximum input size, expected time and space complexity (when documented), and acceptance rate. Prior work has shown that such structural and performance-based features are highly predictive of perceived difficulty \citep{effenberger2019difficulty,skarbalius2021difficulty,wang2024difficulty}.
\end{itemize}

GPT-4o receives only the textual problem descriptions, with all numeric metadata (e.g., explicit constraints table, acceptance rate) removed or paraphrased away so that they cannot be trivially parsed as numbers. This configuration isolates the model's ability to infer difficulty from natural language alone. In contrast, the LightGBM ensemble ingests both the TF--IDF features and the numeric indicators. We split the data into training, validation, and held-out test sets and evaluate both models using accuracy, macro-precision, macro-recall, and macro F1-score.

\subsection{LLM Prompting and Labeling Protocol}
GPT-4o is prompted with a standardized instruction: given a problem description, it must assign one of three labels \{\texttt{Easy}, \texttt{Medium}, \texttt{Hard}\} that best reflects the typical experience of an upper-intermediate competitive programmer on LeetCode. To reduce prompt sensitivity, we manually tune a small set of prompt variants and fix the best-performing one on a validation subset. For the main experiments, each problem is labeled once, without self-consistency sampling or chain-of-thought prompting, mirroring common usage in LLM-as-judge setups \citep{li2024llmjudge,bavaresco2025llmsjudges}.

We log the raw distribution of GPT-4o labels across the three classes, as well as per-class confusion against the ground-truth LeetCode difficulty. This allows us to quantify class-wise biases (e.g., a strong preference for Medium) in addition to overall accuracy.

\subsection{LightGBM Ensemble and Interpretability}
Our structured baseline is a LightGBM ensemble classifier \citep{ke2017lightgbm} trained on the combined textual and numeric features. We perform hyperparameter tuning via grid search on the validation set, optimizing for macro F1-score. Once trained, we compute SHAP values \citep{lundberg2017shap} to assess the relative importance of each feature, focusing particularly on (i) input size constraints and (ii) acceptance rate.

We then visualize global feature importance (mean absolute SHAP values) as well as class-specific SHAP summaries, examining how high values of constraint-related features shift the log-odds toward the Hard label. This analysis provides a contrastive explanation for why the LightGBM ensemble can robustly distinguish Hard problems, while GPT-4o systematically fails to do so.

\subsection{Synthetic Hard Problem Generation}
To probe GPT-4o's internal notion of ``Hard,'' we design a two-stage synthetic experiment. First, we select 21 representative Hard problems from the original dataset, chosen to span different algorithmic themes (e.g., graph algorithms, dynamic programming, advanced data structures). For each seed title, we prompt GPT-4o to generate a new Hard problem that:

\begin{enumerate}
    \item Maintains the original algorithmic core (e.g., remains a graph shortest-path problem),
    \item Introduces large input sizes and tight constraints comparable to typical Hard problems,
    \item And includes a short solution sketch with the expected time and space complexity.
\end{enumerate}

This yields 385 synthetic problems, each with an associated GPT-4o-generated label. We then remove any explicit meta-commentary about difficulty and re-prompt GPT-4o to \emph{re-classify} each synthetic problem using the same judgment protocol as for the real dataset. This self-judgment step mirrors recent work that uses LLMs both to generate and to evaluate content in educational and assessment settings \citep{seo2025educationeval,ji2024fieldtesting}.

\section{Results}

\subsection{Overall Performance}
Table~\ref{tab:results} summarizes the performance of GPT-4o and the LightGBM ensemble on the original dataset of 1{,}825 problems. GPT-4o attains a meager 37.75\% accuracy, while LightGBM achieves 86\%. Macro F1-score exhibits a similar gap. Most misclassifications by GPT-4o involve Hard problems being labeled as Medium or Easy, confirming the label distribution analysis below.

\begin{table}[ht]
\centering
\caption{Performance comparison on the original dataset (1{,}825 problems). Macro-averaged precision, recall, and F1-score are reported over the three difficulty classes.}
\label{tab:results}
\begin{tabular}{lcccc}
\toprule
\textbf{Model} & \textbf{Accuracy} & \textbf{Precision} & \textbf{Recall} & \textbf{F1-Score} \\
\midrule
GPT-4o & 37.75\% & 40.9\% & 31.5\% & 35.6\% \\
LightGBM Ensemble & 86.0\% & 85.2\% & 82.4\% & 83.7\% \\
\bottomrule
\end{tabular}
\end{table}

\subsection{LLM Labeling Distributions}
Beyond aggregated metrics, we inspect GPT-4o's raw label distribution. Among the 385 real Hard problems in the dataset, GPT-4o outputs:
\begin{itemize}
    \item 321 problems (83.38\%) as Easy,
    \item 43 problems (11.17\%) as Medium,
    \item 21 problems (5.45\%) as Hard.
\end{itemize}
Thus, GPT-4o overwhelmingly \emph{downgrades} Hard problems, often directly to Easy. This contrasts with observations that LLMs as judges can sometimes be overly lenient or overly harsh in other domains \citep{li2024llmjudge,bavaresco2025llmsjudges}; here, the bias is highly asymmetric in the direction of underestimating difficulty.

\subsection{Confusion Matrix Analysis for LightGBM}
\begin{figure}[ht]
\centering
\includegraphics[width=0.55\linewidth]{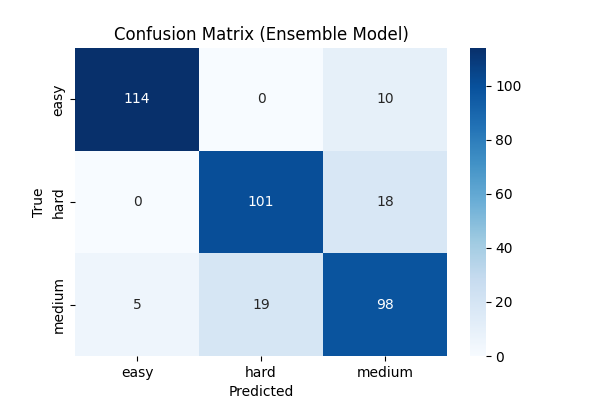}
\caption{Confusion matrix for the LightGBM ensemble on the original dataset, illustrating strong discrimination among Easy, Medium, and Hard problems.}
\label{fig:confmat}
\end{figure}

Figure~\ref{fig:confmat} shows the confusion matrix for the trained LightGBM ensemble. The model accurately classifies the majority of Hard problems, with relatively few misclassifications into Medium and very few into Easy. The Easy and Medium classes are also well separated, with confusion concentrated near problems that intuitively feel ``borderline'' between two levels.

This pattern indicates that the numeric and structural features are highly informative and that the ensemble can exploit them to build a robust decision boundary. In particular, Hard problems often involve stricter time constraints and lower acceptance rates, consistent with prior findings that complexity and performance indicators correlate with experienced difficulty \citep{effenberger2019difficulty,wang2024difficulty}.

\subsection{Feature Importance via SHAP}
\begin{figure}[ht]
\centering
\includegraphics[width=0.6\linewidth]{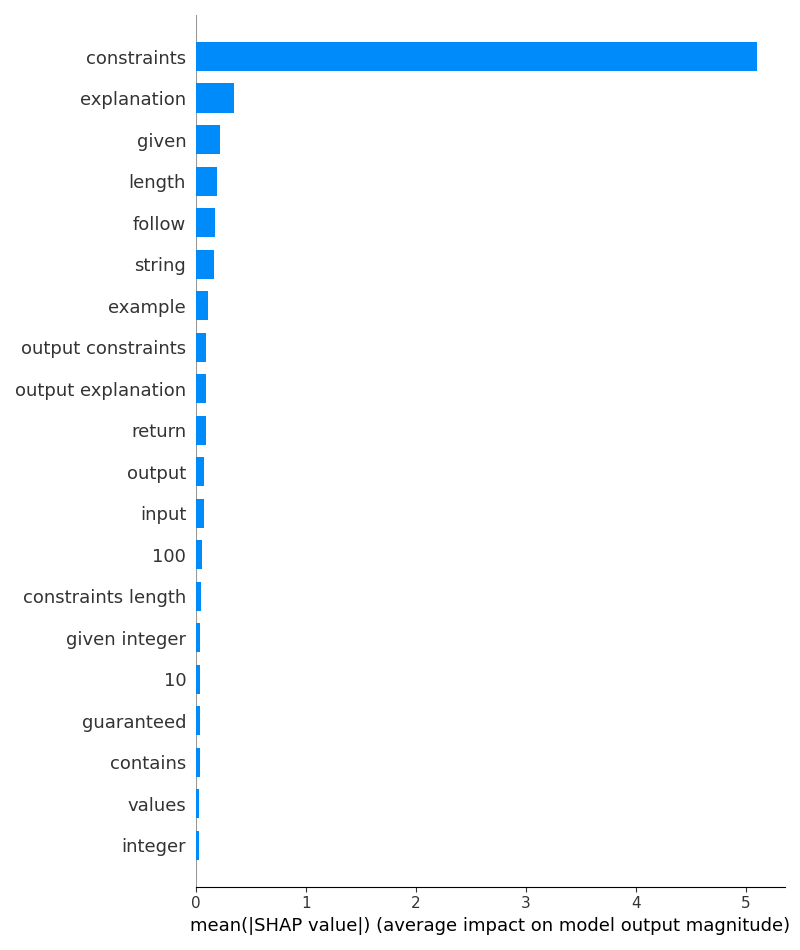}
\caption{Global SHAP bar plot indicating that constraint-related features (e.g., maximum input size, acceptance rate) dominate LightGBM's classification of Hard tasks.}
\label{fig:shap}
\end{figure}

To understand why LightGBM succeeds where GPT-4o fails, we examine global SHAP feature importance for the ensemble. Figure~\ref{fig:shap} reports the mean absolute SHAP value per feature. Numeric constraints---especially maximum input size and acceptance rate---are among the most influential features for classifying Hard problems. The SHAP value distributions show that:

\begin{itemize}
    \item Larger input size limits consistently push predictions toward the Hard class.
    \item Lower acceptance rates (i.e., fewer successful submissions) are strongly associated with Hard labels.
    \item Certain TF--IDF features corresponding to algorithmic keywords (e.g., ``segment tree'', ``max flow'') also contribute, but less strongly than the numeric constraints.
\end{itemize}

These findings align with general observations that SHAP can reveal meaningful feature attributions in tree-based models \citep{lundberg2017shap}. In our context, they highlight a discrepancy: the numeric cues that LightGBM relies on heavily are either removed or implicitly described in the GPT-4o prompts, and the LLM does not reconstruct them accurately from the remaining text.

\subsection{Synthetic Hard Problem Generation}
We now turn to the synthetic setting. GPT-4o is instructed to generate 385 new Hard problems based on 21 seed Hard titles, each accompanied by a short solution sketch and explicit constraints. When asked to label these problems:

\begin{itemize}
    \item 384 problems (99.74\%) are labeled as Medium,
    \item 1 problem (0.26\%) is labeled as Hard,
    \item 0 problems are labeled as Easy.
\end{itemize}

Thus, GPT-4o's self-generated problems, written under explicit Hard instructions and with constraints that it itself declares as large, are nearly always classified as Medium. This behavior contradicts its pattern on real Hard problems, which are mostly labeled as Easy. The resulting picture is that GPT-4o's internal boundary between Easy, Medium, and Hard is not only misaligned with LeetCode's ground truth, but also \emph{unstable} across real and synthetic domains.

Such instability echoes broader concerns from the LLM-as-judge literature, where calibration and robustness of judgments across datasets and task formulations are known challenges \citep{li2024llmjudge,bavaresco2025llmsjudges}. Here, we see a concrete manifestation in the context of difficulty assessment for competitive programming.

\section{Discussion}
Our experiments expose two intertwined issues in GPT-4o's difficulty judgments.

\paragraph{Systematic underestimation of real Hard problems.}
The extreme downgrading of real Hard problems (83\% labeled Easy) suggests that GPT-4o's semantic understanding of problem text is insufficient to capture the algorithmic and numeric nuances that define Hard difficulties. Prior work on difficulty estimation \citep{zhao2018topicsdifficulty,effenberger2019difficulty,skarbalius2021difficulty,wang2024difficulty} emphasizes that structural complexity and performance statistics are critical, and our SHAP analysis confirms that numeric constraint features drive LightGBM's success. GPT-4o, deprived of explicit numeric metadata, appears to treat many Hard problems as ordinary variants of medium-level algorithmic exercises.

\paragraph{Synthetic Hard problems collapse into Medium.}
In contrast, GPT-4o-generated Hard problems are almost universally classified as Medium when judged by the same model. This behavior hints that the model's notion of a ``Hard'' label is anchored to a narrow subset of textual cues that it rarely emits even when instructed, and that these cues differ qualitatively from the structure of real Hard problems. The phenomenon resembles discrepancies between LLM-generated content and authentic human data in other evaluation settings \citep{bavaresco2025llmsjudges,li2024llmjudge}. It also raises questions about the use of synthetic Hard problems in training or evaluating difficulty-aware systems: if the synthetic tasks lack the stringent constraints typical of real Hard problems, they may systematically underestimate the true difficulty distribution.

\paragraph{Implications for educational platforms and RL pipelines.}
For educational platforms and online judges, mis-calibrated difficulty labels can distort learners' experience and progression. A problem labeled Easy but felt as Hard may discourage novices, while a genuinely Hard problem mislabeled as Medium can create frustration and misrepresent mastery. Recent studies on using LLMs to evaluate and provide feedback in education \citep{seo2025educationeval,woodrow2025dpo} highlight both the promise and the risks of delegating evaluation to black-box models. Our results suggest that, at least for competitive programming difficulty, relying solely on LLM judgments is not advisable.

Similarly, in reinforcement learning from human feedback and related alignment frameworks \citep{christiano2017deep}, difficulty-aware reward models are sometimes proposed to scale up training and curriculum design. If LLMs are used as proxies for judges in such pipelines, the type of misalignment we observe here could bias reward signals and hinder learning progress.

\paragraph{The value of interpretable numeric models.}
Our LightGBM baseline, though relatively simple, offers a counterpoint: its decisions are both accurate and interpretable. SHAP analysis reveals that intuitive features---input sizes and acceptance rates---drive the Hard classification, providing a transparent rationale. This suggests a promising hybrid strategy: use LLMs to enrich textual representations and generate candidate features, but rely on interpretable numeric models to anchor difficulty judgments in measurable constraints. This is consistent with broader calls for hybrid, human-aligned evaluation systems \citep{li2024llmjudge,alkhuzaey2024review}.

\section{Conclusion and Future Work}
We investigated GPT-4o as a difficulty judge for LeetCode-style programming problems and found major discrepancies between its judgments and those of an interpretable LightGBM ensemble trained on numeric and textual features. GPT-4o drastically underestimates real Hard problems, frequently labeling them as Easy, while its own synthetic Hard problems are almost always classified as Medium. In contrast, LightGBM, supported by SHAP-based interpretability, shows robust performance and highlights the primacy of numeric constraints in defining Hard difficulty.

Our findings contribute to the growing literature on LLMs-as-judges by providing a case study in a highly structured domain, and suggest several avenues for future work:

\begin{itemize}
    \item \textbf{Constraint-aware prompting.} We plan to explore prompts that explicitly foreground numeric details, including structured tables of constraints, to test whether GPT-4o can more reliably internalize and use them for difficulty assessment.
    \item \textbf{Hybrid models.} Building on recent multimodal difficulty models \citep{wang2024difficulty}, we aim to combine LLM-derived embeddings of problem text and solution code with interpretable numeric features, potentially using LightGBM or related tree ensembles as the final judge.
    \item \textbf{Meta-evaluation protocols.} Inspired by work on meta-evaluating LLM judges \citep{li2024llmjudge,bavaresco2025llmsjudges}, we seek to design benchmark suites and agreement metrics specifically tailored to difficulty assessment, including human studies on perceived difficulty and fairness.
\end{itemize}

By bridging the gap between text-based reasoning and robust numeric analysis, we hope to pave the way toward trustworthy AI judges for competitive programming and beyond.


\end{document}

%% file: math_commands.tex
\usepackage{amsmath,amsfonts,bm}

\def\eqref#1{equation~\ref{#1}}

\def\1{\bm{1}}

\DeclareMathAlphabet{\mathsfit}{\encodingdefault}{\sfdefault}{m}{sl}
\SetMathAlphabet{\mathsfit}{bold}{\encodingdefault}{\sfdefault}{bx}{n}

